%% file: main_seperate.tex
\documentclass[11pt, a4paper, logo, internal]{dm}
\usepackage{graphicx}%
\usepackage{multirow}%
\usepackage{amsmath,amssymb,amsfonts}%
\usepackage{amsthm}%
\usepackage{mathrsfs}%
\usepackage[title]{appendix}%
\usepackage{xcolor}%
\usepackage{textcomp}%
\usepackage{manyfoot}%
\usepackage{booktabs}%
\usepackage{algorithm}%
\usepackage{algorithmicx}%
\usepackage{algpseudocode}%
\usepackage{listings}%
\usepackage{hyperref}
\usepackage{lineno}

\theoremstyle{thmstyleone}%
\theoremstyle{thmstyletwo}%

\theoremstyle{thmstylethree}%
\input{defns}

\def\method{Hygieia}

\title{A Versatile AI Agent for Rare Disease Diagnosis and Risk Gene Prioritization}

\begin{document}




\author[1,2,3]{Tianyu Liu}
\author[2]{Wangjie Zheng}
\author[4]{Rui Yang}
\author[5,6,17]{Benny Kai Guo Loo}
\author[6,17]{Hui Zhang}
\author[6,17]{Jeffries Lauran}
\author[2]{Jianlei Gu}
\author[7]{Botao Yu}
\author[8,9]{Weihao Xuan}
\author[10]{Kexin Huang}
\author[4,11,12]{Nan Liu}
\author[13]{James Zou}
\author[14,15]{Yonghui Jiang}
\author[1,15,16,*]{Hua Xu}
\author[1,2,14,15,16,*]{Hongyu Zhao}

\affil[1]{Interdepartmental Program in Computational Biology and Bioinformatics, Yale University}

\affil[2]{Department of Biostatistics, Yale University}

\affil[3]{Broad Institute of MIT and Harvard}

\affil[4]{Center for Biomedical Data Science, Duke-NUS Medical School}

\affil[5]{Sport and Exercise Medicine Service, KK Women's and Children's Hospital}

\affil[6]{Training Program, Duke-NUS Medical School}

\affil[7]{Department of Computer Science and Engineering, The Ohio State University}

\affil[8]{Department of Complexity Science and Engineering, The University of Tokyo}

\affil[9]{Center for Advanced Intelligence Project, RIKEN}

\affil[10]{Phylo}

\affil[11]{NUS Artificial Intelligence Institute, National University of Singapore}

\affil[12]{Department of Biostatistics and Bioinformatics, Duke University}

\affil[13]{Department of Biomedical Data Science, Stanford University}

\affil[14]{Department of Genetics, Yale University}

\affil[15]{Wu Tsai Institute, Yale University}

\affil[16]{Department of Biomedical Informatics and Data Science, Yale University}

\affil[17]{These authors contributed equally to this work as human experts.}

\affil[*]{Corresponding authors.}

\begin{abstract}
Accurate and timely diagnosis is essential for effective treatment, particularly in the context of rare diseases. However, current diagnostic workflows often lead to prolonged assessment times and low accuracy. To address these limitations, we introduce Hygieia, a multi-modal AI agent system designed to support precision disease diagnosis by integrating diverse data sources, including phenotypic features, genetic profiles, and clinical records. Hygieia features a router-based and knowledge-enhanced framework that mitigates hallucination and tailors diagnostic strategies to different disease categories. Notably, it prioritizes risk-related genomic factors for rare diseases and provides confidence scores to assist clinical decision-making. We conducted a comprehensive evaluation demonstrating that Hygieia achieves state-of-the-art performance across multiple diagnostic benchmarks. In collaboration with clinical experts from Yale School of Medicine and Duke-NUS Medical School, we further validated its practical utility by showing (1) Hygieia’s superior diagnostic performance compared to physicians with an improvement from 12\%-60\% and (2) its effectiveness in assisting clinicians with medical records for handling real-world cases. Our findings indicate that Hygieia not only enhances diagnostic accuracy and interpretability but also significantly reduces clinician workload, highlighting its potential as a valuable tool in clinical decision support systems.
\end{abstract}

\keywords{AI Agent, Disease Diagnosis, Rare Disease, Risk Gene Prioritization, Clinical System}

\maketitle

\section{Introduction}
\input{section_folder/introduction}
\section{Results}
\input{section_folder/results}

\section{Discussion}
\input{section_folder/discussion}

\section{Methods}
\input{section_folder/methods}

\bibliographystyle{unsrt}
\bibliography{sn-bibliography}

\newpage

\appendix
\input{section_folder/Appendix}

\end{document}

%% file: defns.tex
\usepackage{mathtools}
\usepackage[dvipsnames]{xcolor}
\usepackage[numbers, sort&compress, round]{natbib}
\usepackage{booktabs}
\usepackage{graphicx}
\usepackage{xfrac}
\usepackage{bbm}
\usepackage{changes}
\usepackage{rotating}
\usepackage{wrapfig}

\usepackage[most]{tcolorbox}
\usepackage{xparse}
\usepackage{adjustbox}
\usepackage{xspace}
\usepackage{changepage}
\usepackage{enumitem}
\usepackage{pifont}
\usepackage{ulem}
\usepackage{tocloft}
\usepackage[toc]{multitoc}
\usepackage{etoc}
\usepackage{dsfont}
\usepackage{dm-colors}
\usepackage{multirow}
\usepackage{minted}
\usepackage{caption}
\usepackage{soul}
\usepackage{float}
\usepackage{svg}
\usepackage{adjustbox}
\usepackage{setspace}
\svgsetup{inkscapelatex=false}
\usepackage{silence}  
\newcommand{\hide}[1]{}

\newcommand{\method}{SpatialAgent\,}

\newcommand{\cmark}{\ding{51}}
\newcommand{\xmark}{\ding{55}}

\pdftrailerid{redacted}

%% file: section_folder/introduction.tex
Rare diseases are defined as conditions affecting fewer than 1 in 2,000 individuals, affecting over 300 million patients worldwide \cite{valdez2016public,nguengang2020estimating,paul2013hope,jonker2024access,zhao2025agentic}. Moreover, the diagnosis and test recommendations for rare diseases are very challenging. Diagnosing rare diseases based on conventional medical methods typically takes 4 to 5 years \cite{ghosh2025artificial}, which is known as ``diagnostic odyssey''. One reason is that the phenotypes of rare diseases can sometimes be difficult to distinguish directly from common diseases. This also increases the likelihood of misdiagnosis and mistreatment by physicians \cite{dong2020misdiagnosis}. 

To overcome the challenges mentioned above, researchers have begun collecting genotypes, phenotypes, and diagnosis plans for rare diseases to enrich our understanding of these diseases and to design a better diagnosis and treatment plan. These multi-modal datasets provide valuable research materials for disease diagnosis. Moreover, data-driven solutions, such as disease diagnosis methods based on data mining, deep learning, and advanced artificial intelligence (AI), have also garnered significant attention recently \cite{lee2022deep}. Experts can train a model using the aforementioned diagnostic data to predict diseases or prioritize disease risk genes, thereby improving diagnostic accuracy. However, such models often face shortcomings in terms of generalization and deployment \cite{rossi2025meta}. To enhance the model's versatility and accessibility for physicians, rare disease diagnostic tools based on Foundation Models (such as Large Language Models (LLMs) \cite{thirunavukarasu2023large} and Visual Language Models (VLMS) \cite{liu2025visual}) have also been developed. LLMs are pre-trained with a large-scale text corpus and can generalize into different tasks in natural language processing (NLP) with techniques of post-training. LLMs can process electronic health records (EHRs) from patients and make diagnoses accordingly \cite{sarker2024natural,liu2024geneverse}, and multiple LLMs with different roles can also work together as an agent system \cite{tran2025multi,du2025accelerating}. Such an AI agent can make diagnoses by simulating the real scenarios, leveraging prior knowledge, and providing recommendations and suggestions for physicians as medical AI assistants \cite{liu2025spemo,liu2025teampath}. 

Several AI-based tools have been developed for rare disease diagnosis. For example, a general baseline for rare disease diagnosis will be prompting LLMs discussed in RareBench \cite{chen2024rarebench}, and RareArena \cite{chen2026rarearena}. Researchers also consider developing AI agents for medical usage based on techniques such as knowledge-enhanced retrieval \cite{wang2025visualrag} as well as multi-agent communication \cite{dhatterwal2023multi,chen2025enhancing}. Although these models are interesting in design and have some clinical significance, their shortcomings are still quite evident. First, even advanced AI-based models used for rare disease diagnosis cannot distinguish between common and rare diseases, which is the basic step to avoid misdiagnosis (Supplementary Figure \ref{supfig:deeprare_case} (a)). This finding severely limits the applicability of rare disease diagnostic models. Second, due to the randomness that exists in the model training and inference, these models might not give consistent outputs based on the same input with different random seeds, which is also harmful for the trustworthy output (Supplementary Figure \ref{supfig:deeprare_case} (b)). Third, current AI methods only focus on diagnosis, but lack the necessary steps and capacities for result interpretation and discovery of important causal genes of rare diseases. Finally, current studies \cite{lee2022deep, zhao2025agentic, liu2026vc} do not directly address how these AI models can be applied in diagnostic scenarios and collaborate with physicians. Moreover, recent studies have shown that over 80\% of rare diseases are influenced by genetic factors and can be passed on to the next generation \cite{zhao2025agentic}, but how these genetic factors are incorporated into diagnosis and how to infer disease-related genes based on clinical presentation have not been well researched in AI-based methods. Therefore, there is a critical need to design an AI model that can simultaneously diagnose different types of diseases and provide explanations for the decision. 

In this manuscript, we introduce \method{}, which is an AI agent for disease diagnosis and interpretation. Our model breaks down the diagnostic process into multiple stages, first determining the disease type, then designing distinct diagnostic approaches for common and rare diseases. Diagnosing common diseases is based on prompting LLMs. Meanwhile, due to the complexity of rare disease diagnosis, our agent utilizes multiple tools (such as website searching as well as patient retrieval) to leverage prior knowledge and make a decision. We have two innovations in the design of this agent. First, to resolve the inconsistency, our agent has a verifier to monitor the outputs of the main body of \method{} and ensure the results converge. Second, to improve the transparency of using AI agents for making clinical decisions, we implement a framework with a reasoning trajectory and confidence estimation to help users understand and trust this workflow. \method{} can accept multiple modalities or types of data as inputs, such as phenotype information, gene information, medical history, and other information. We also provided a table in Appendix \ref{appendix:agentcompare} to distinguish \method{} versus other AI agents focusing on (rare) disease diagnosis, and the unique components (case router, confidence estimation, and multi-task capacity) of \method{} further enhance its novelty.

In general, \method{} can interpret diagnosis results with trackable reasoning paths and prioritize disease-associated risk genes, providing more informative feedback as references to help physicians make decisions and recommendations. \method{} can also be integrated as a skill to improve the harness of frontier AI Agents in the diagnosis of rare diseases. We invite genetic physicians with verification from Yale School of Medicine and DUKE-NUS Medical School to evaluate the contribution of \method{} as an effective medical AI assistant, and explore new directions for the diagnosis of rare diseases at the age of AI and digital health.

%% file: section_folder/results.tex
\begin{figure}
    \centering
    \includegraphics[width=1\linewidth]{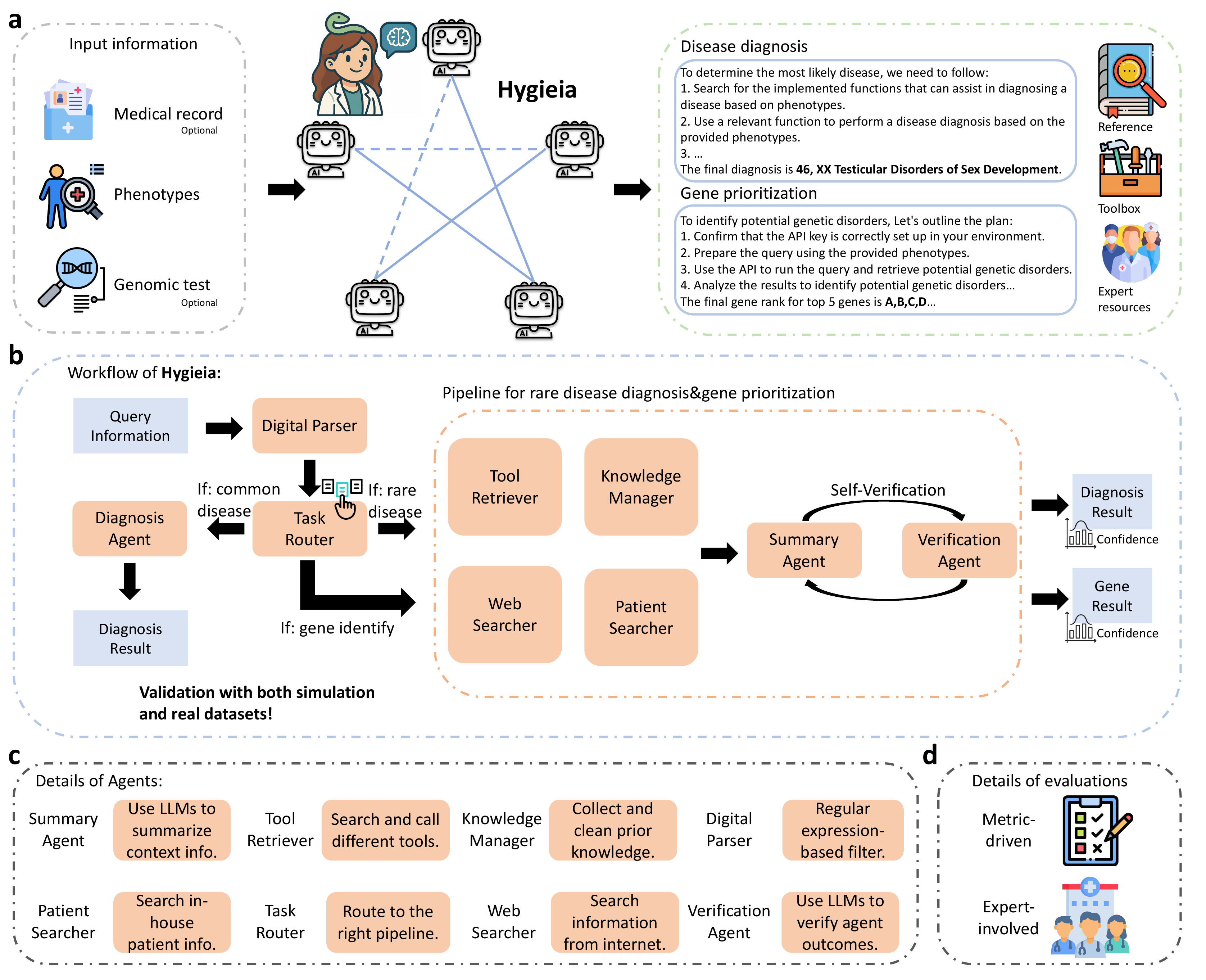}
    \caption{Overall pipeline of \method{}. (a) Here we showcase how \method{} can help physicians and clinicians working on two important problems in rare disease analysis, including diagnosis and risk gene prioritization. (b) The workflow of the AI Agent pipeline. We have multiple components, first routing the AI agents with correct models based on inferred disease type, and then providing diagnosis outcomes as well as confidence. (c) We provide the detailed information of each component in our AI agent. (d) Our evaluation criteria, including numerical evaluation and human evaluation, to support mimicking the scenario of clinical application.}
    \label{fig:figoverall}
\end{figure}

\textbf{Method overview.} For each patient, we collect the annotations from phenotypes, medical records, and genetic test records as the input of \method{}, and then make a decision based on the prior knowledge in medical research retrieved from the internet and databases. To initiate the workflow, \method{} will first compute the probability of disease type based on a router, and then determine the most suitable pipeline for diagnosing either common or rare diseases. The pipeline used for rare disease diagnosis is more complex and involves several agents, such as a knowledge-retrieval agent, an information-extraction agent, a summary agent, and a verification agent. \method{} also infers the risk gene based on patient-level phenotypes or medical records, as an extra function for medical geneticists. Finally, we provide the confidence level of the model output through a majority voting approach. The input data types, workflow, and application scenarios of \method{} are summarized in Figure \ref{fig:figoverall}. Our benchmarking datasets include MyGene2 \cite{rodrigues2022variant}, four splits from RareBench \cite{chen2024rarebench} (RAMEDIS, MME, HMS, and LIRICCAL), RareArena \cite{chen2026rarearena}, and in-house data from Yale School of Medicine (YSM) and Yale New Haven Hospitals (YNHH) used for the comparison between human experts and \method{}.

\begin{figure}
    \centering
    \includegraphics[width=1\linewidth]{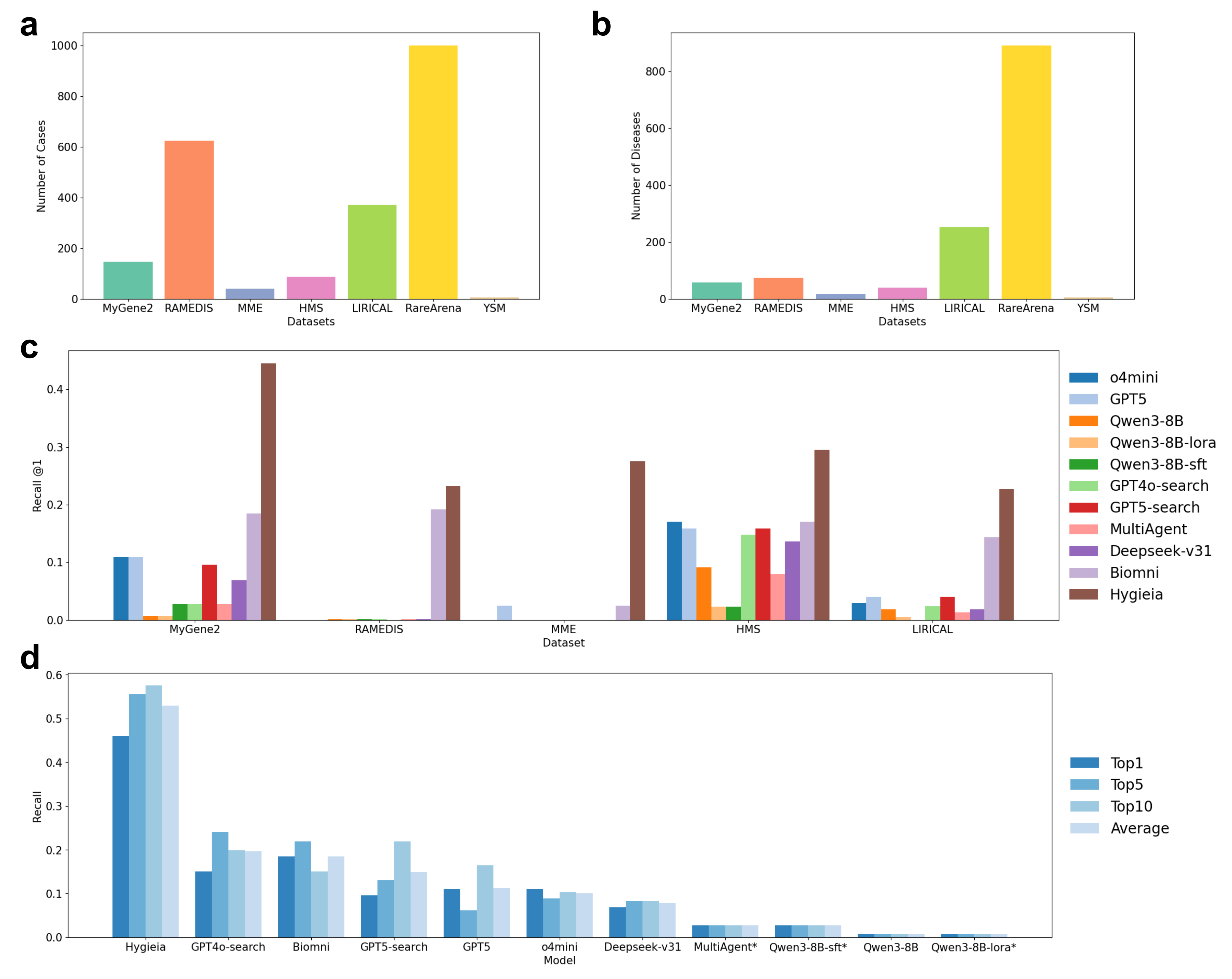}
    \caption{Benchmarking results for \method{} in rare disease diagnosis. (a) Number of cases in our testing datasets. (b) Number of diseases in our testing datasets. (c) Top1 recall rate across different datasets. (d) Comparisons of different models with different recall rates in MyGene2.}
    \label{fig:rarebench1}
\end{figure}

\textbf{\method{} serves as a strong medical agent for disease diagnosis.} We first demonstrate the strong capacity of \method{} serving as a virtual physician, supported by validations conducted from various clinical datasets with different sources. The accuracy of disease classification is much higher than using random guess or prompting LLMs (Supplementary Figure \ref{supfig:diseasrouter} (a)), which shows the advantages of having a router to set up a simpler pipeline for common disease diagnosis. Our classifier also predicts accurately for the correct disease types not only in our held-out testing split, but also external validation datasets (MyGene2 and RareBench), shown in Supplementary Figure \ref{supfig:diseasrouter} (b). We also visualize the distribution of embeddings with Uniform Manifold Approximation and Projection (UMAP) \cite{mcinnes2018umap} in Supplementary Figure \ref{supfig:diseasrouter} (c), colored by disease types, where we see a clear difference between samples with rare diseases and samples with common diseases, which works as an explanation for our contribution. The diagnosis of rare diseases is more difficult, and we collect seven datasets in our evaluation pipeline. Figure \ref{fig:rarebench1} (a) shows the distribution of case numbers across all datasets, while Figure \ref{fig:rarebench1} (b) shows the unique number of diseases in different datasets. These figures show that our selected datasets have obvious differences in data distribution, which helps us simulate the clinical usage in the real-world setting. Among these datasets, RAMEDIS, MME, HMS, and LIRICAL are extracted from RareBench, which is a public benchmark framework for evaluating the performance of models for rare disease diagnosis. MyGene2 contains family-level information for patients with rare diseases, which is also publicly available. RareArena is a newly collected dataset for evaluating LLMs' performances in rare disease diagnosis. The YSM dataset contains five in-house samples as a part of Undiagnosed Diseases Network (UDN) \cite{ramoni2017undiagnosed}, which covers diseases that are hard to diagnose. This dataset is not directly accessible to public researchers to protect personal information. 

Regarding baseline methods, we also consider a group with strong diversity. Our baselines include LLMs with/without reasoning and searching capacities, agents for biomedical research, and LLMs finetuned with simulated patient-level data from \cite{alsentzer2025few} with the open-source Qwen3 model \cite{yang2025qwen3}. Details of our baseline methods can be found in the Methods section.

During evaluation, we test the Recall rate by comparing the observed diseases with the predicted diseases from different methods. Figure \ref{fig:rarebench1} (c) shows that \method{} outperforms various baseline methods across datasets from different resources. Fine-tuning LLMs for disease diagnosis also does not have strong generalization ability, and thus cannot outperform most of the agent-based solutions. Moreover, since some rare diseases have similar phenotypes or can be treated similarly \cite{griggs2009clinical}, we also test if increasing the size of predicted targets (e.g., top 5 and top 10 diseases) can improve the Recall rate. Figure \ref{fig:rarebench1} (d) shows our results based on the MyGene2 dataset, where \method{} had a higher recall rate when we increased the testing size. We report the benchmarking results based on a randomly sampled subset from RareArena in Supplementary Figure \ref{supfig:rarearena_info}, where \method{} still presents a leading performance even under the evaluation settings with an advanced dataset. At the same time, not all methods exhibit this phenomenon, which is counterintuitive. This might indicate that the approach of some models taken to clinical decision-making may still be primarily based on guesswork, with reasoning playing a secondary role. However, by integrating the prior knowledge and introducing the self-reflection design, \method{} achieves a better result in diagnosing rare diseases.

We also test the reliability of estimated confidence, shown in Supplementary Figure \ref{supfig:conf_est} (a). Our expectation is that \method{} has higher confidence for questions with correct answers. Our results demonstrate that 1. the confidence proposed by \method{} is reliable, as the answer group with higher confidence also has a higher recall rate, and 2. other confidence estimation methods are worse than the current design, shown in Supplementary Figure \ref{supfig:conf_est} (b) as the rest of three methods cannot produce significant differences between these two groups. We also test the robustness of \method{} by using three different random seeds for querying, and based on Supplementary Figure \ref{supfig:robustness_check}, we do not observe a large difference across the outcomes with three different runs.

\textbf{Case study of \method{} for disease diagnosis shows its unique and informative decision-making process.} To understand the decision-making process of \method{} and make an in-depth comparison with other baseline methods, we provide a case study with MyGene2. In this sample, we provide phenotypes and detected genes as inputs; the diagnosis is \textit{Distal arthrogryposis, type 10}. The full reasoning path of \method{} can be found in Supplementary Figure \ref{supfig:raredis_example} (a).

\begin{figure}
    \centering
    \includegraphics[width=1\linewidth]{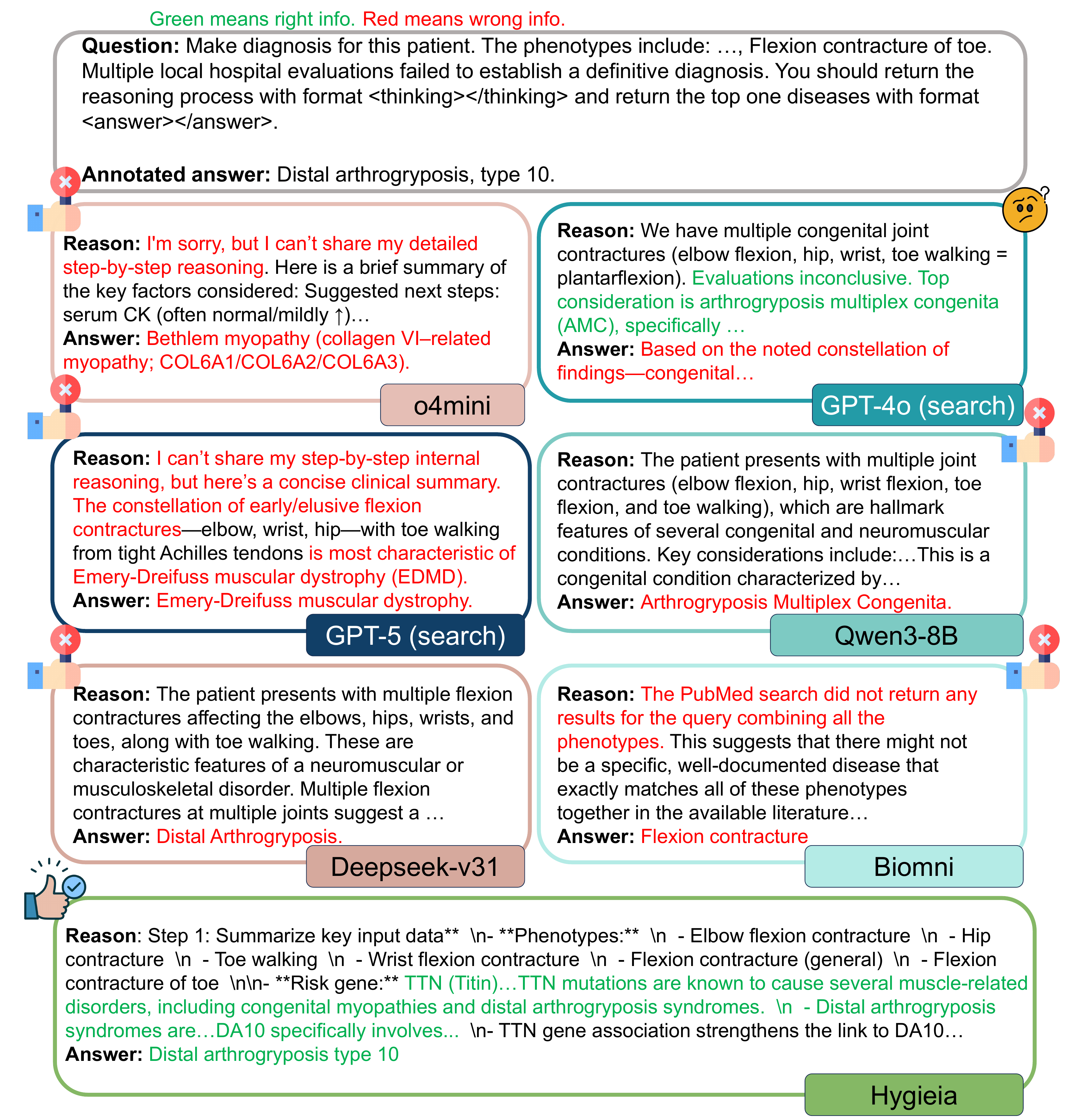}
    \caption{Case study of \method{} and other baselines in disease diagnosis. We mask some phenotypes to protect personal information.}
    \label{fig:examples_rarebench}
\end{figure}

Based on Figure \ref{fig:examples_rarebench} and among all evaluated methods, \method{} is the only model that correctly made the definitive diagnosis \textit{Distal arthrogryposis, type 10}, fully matching both the annotated answer and the genetic etiology (\textit{TTN} mutation). In contrast, alternative baselines either provide incorrect diagnoses (e.g., \textit{Bethlem myopathy, Emery–Dreifuss muscular dystrophy, Flexion contracture, Arthrogryposis multiplex congenita}) or produce overly broad, nonspecific conclusions. This demonstrates that \method{} not only retrieves the correct disease category but is capable of fine‐grained subtype resolution, as an essential requirement in precision medicine.

Unlike other models, \method{} has a multi‐step reasoning process that integrates phenotypic patterns, risk gene associations, and syndrome specificity. This demonstrates a higher level of biomedical causal interpretability. While several baselines refuse to show internal reasoning or default to vague clinical summaries, \method{} transparently links phenotype, genotype, and nosology, exhibiting a cognitively valid chain of inference.

This specific case intentionally includes multiple phenotypes designed to confound rule‐based or pattern‐matching systems. Models like o4-mini, GPT-4o, and Qwen3-8B fail to integrate the constellation of findings, instead overfitting to a single clinical feature (e.g., \textit{toe walking, wrist contracture}) and outputting unrelated diagnoses. \method{}, however, successfully recognizes that the multi‐joint congenital contracture pattern represents a diagnostic signature of distal arthrogryposis, showing resilience to feature redundancy and phenotypic noise.

We also select a challenging example from the YNHH to demonstrate \method{}'s potential for solving complex problems. According to Supplementary Figure \ref{supfig:raredis_diff_example}, \method{} successfully makes the diagnosis of Kabuki Syndrome, including subtypes related to genes based on the iteration steps. Based on the prevalence, only 1 in 32,000 births results in Kabuki syndrome \cite{bogershausen2013unmasking}, making it a very rare and challenging case, and demonstrating the potential of \method{} for handling difficult samples and very rare cases.

Where competing models either offer no next steps or suggest non‐specific investigations, \method{} explicitly ties the diagnosis to a recognized molecular driver and its associated disease spectrum. This strengthens the translational value of its output, enabling downstream steps such as confirmatory genetic testing, family counseling, and prognosis stratification. The output is not merely a label, but clinically operational knowledge, surpassing the diagnostic passivity of baseline systems. Taken together, the evidence indicates that \method{} demonstrates a substantially higher standard of diagnostic precision, biomedical reasoning depth, and clinical applicability compared to both traditional LLM baselines and search‐augmented systems. Its ability to unify phenotypic complexity with molecular knowledge exemplifies the next generation of AI‐assisted medical decision systems, thereby positioning \method{} as the most reliable and clinically aligned model in this evaluation.

\textbf{\method{} successfully prioritizes genes with higher disease risks from individual-level data.} To enhance diagnostic interpretability and confidence while providing additional therapeutic insights, modern medicine often aims to identify causative factors \cite{sanchez2022causal}. In rare disease diagnosis, some physicians may recommend exome- or whole-genome sequencing or targeted gene sequencing to identify disease-causing variants, enabling more reliable conclusions \cite{boycott2013rare}. Therefore, determining how to provide patients with a list of potential genes for sequencing represents a critical task in rare disease diagnosis. However, few AI agent frameworks have considered this task. Our analysis indicates that disease diagnosis shares similarities with risk factor prioritization, suggesting it can be addressed using a unified framework. Here, our input data are still phenotypes or EHR data, while the output will be a gene or a list of genes. Our selected baselines are similar to the candidates used for evaluating disease diagnosis functions. Details can be found in the Methods section.

\begin{figure}
    \centering
    \includegraphics[width=0.9\linewidth]{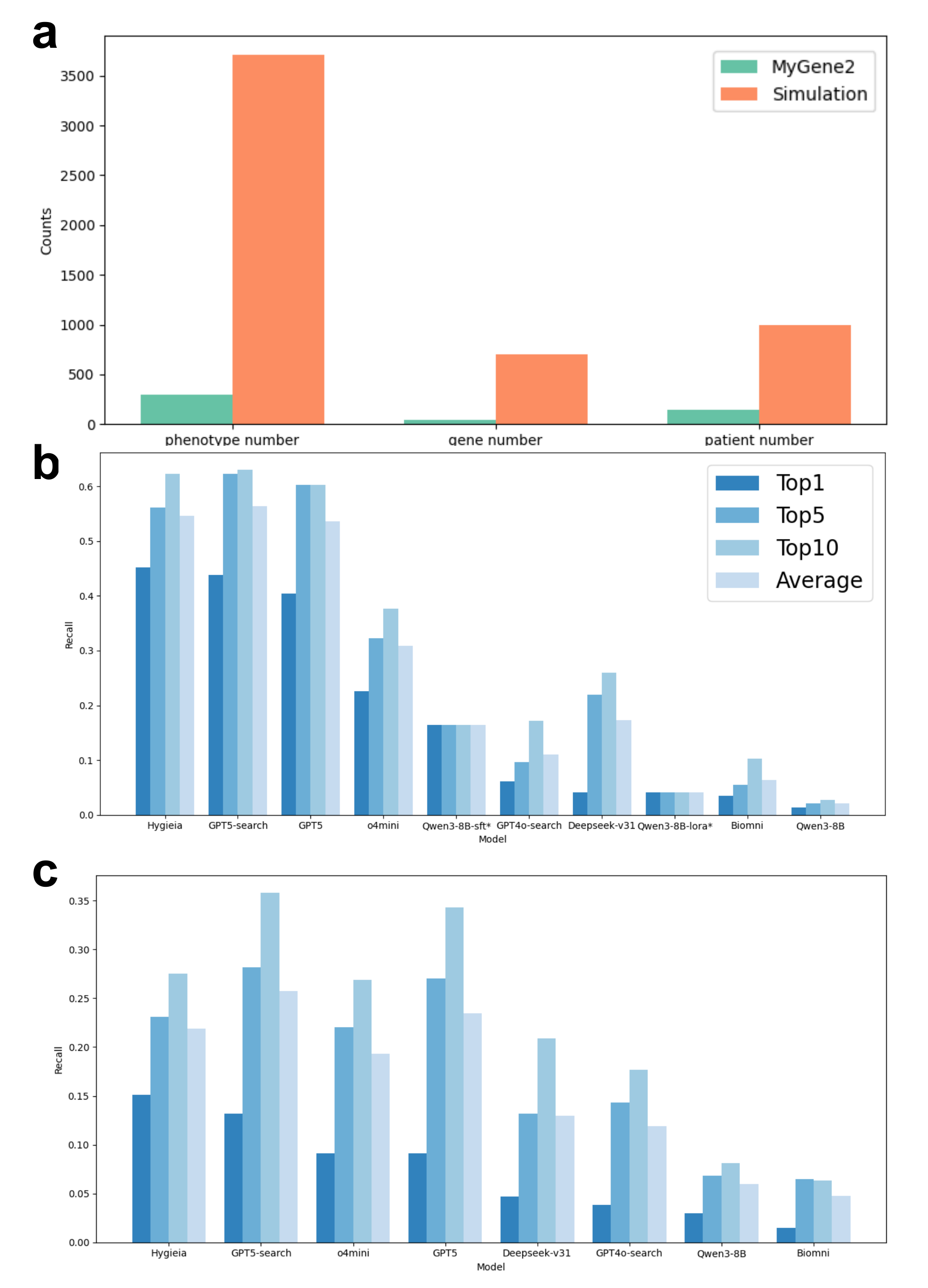}
    \caption{Benchmarking results for \method{} in risk gene prioritization. (a) Statistics in our testing datasets. (b) Top1 recall rate across different datasets. (c) Comparisons of different models with different recall rates in MyGene2. }
    \label{fig:generank_result}
\end{figure}

Due to the scarcity of datasets containing both phenotypes and true disease-causing genes, this section employs MyGene2 and simulation data provided by SHEPHERD \cite{alsentzer2025few} for model evaluation. The statistics of selected datasets are summarized in Figure \ref{fig:generank_result} (a). The scale of simulation data is larger than MyGene2, and thus our assessment also took various scenarios into account. We still computed the Recall rate based on gene lists of different sizes and observed gene labels. Figures \ref{fig:generank_result} (b) and (c) show that \method{} has a high recall rate, especially under the top 1 setting versus other baselines. However, as we increase the pool of candidates, there is a diminishing advantage of \method{}. When comparing the recall rate of this task to disease diagnosis, we observe that gene prioritization is a relatively simpler task. Consequently, as the pool of candidates expands, the benefits of employing agents (e.g., additional verifiers) diminish proportionally. Considering that the more genes that need to be tested, the higher the cost for patients, our approach balances accuracy and expense. We also compared the costs of \method{} and GPT-5-search, revealing that \method{} holds an advantage in token consumption as well, shown in Supplementary Figure \ref{supfig:costanalysis}. In Figure \ref{fig:generank_result} (c), since we can create a training dataset from the large simulation data, we can also create an oracle model (Qwen3-8B-sft, score is 0.724). However, the performance of this model for recommending based on MyGene2 is poor, suggesting that AI Agents have better generalization ability than traditional SFT approaches.

\textbf{Case study of \method{} for gene ranking shows its unique and informative decision-making process.} To understand the decision-making process of \method{} and make an in-depth comparison versus other baseline methods, we provided a case study with one sample from MyGene2, but for risk gene prioritization. In this sample, we provided phenotypes as inputs; the observed risk gene is \textit{NALCN}. The full reasoning path of \method{} can be found in Supplementary Figure \ref{supfig:raredis_example} (b).

\begin{figure}
    \centering
    \includegraphics[width=1\linewidth]{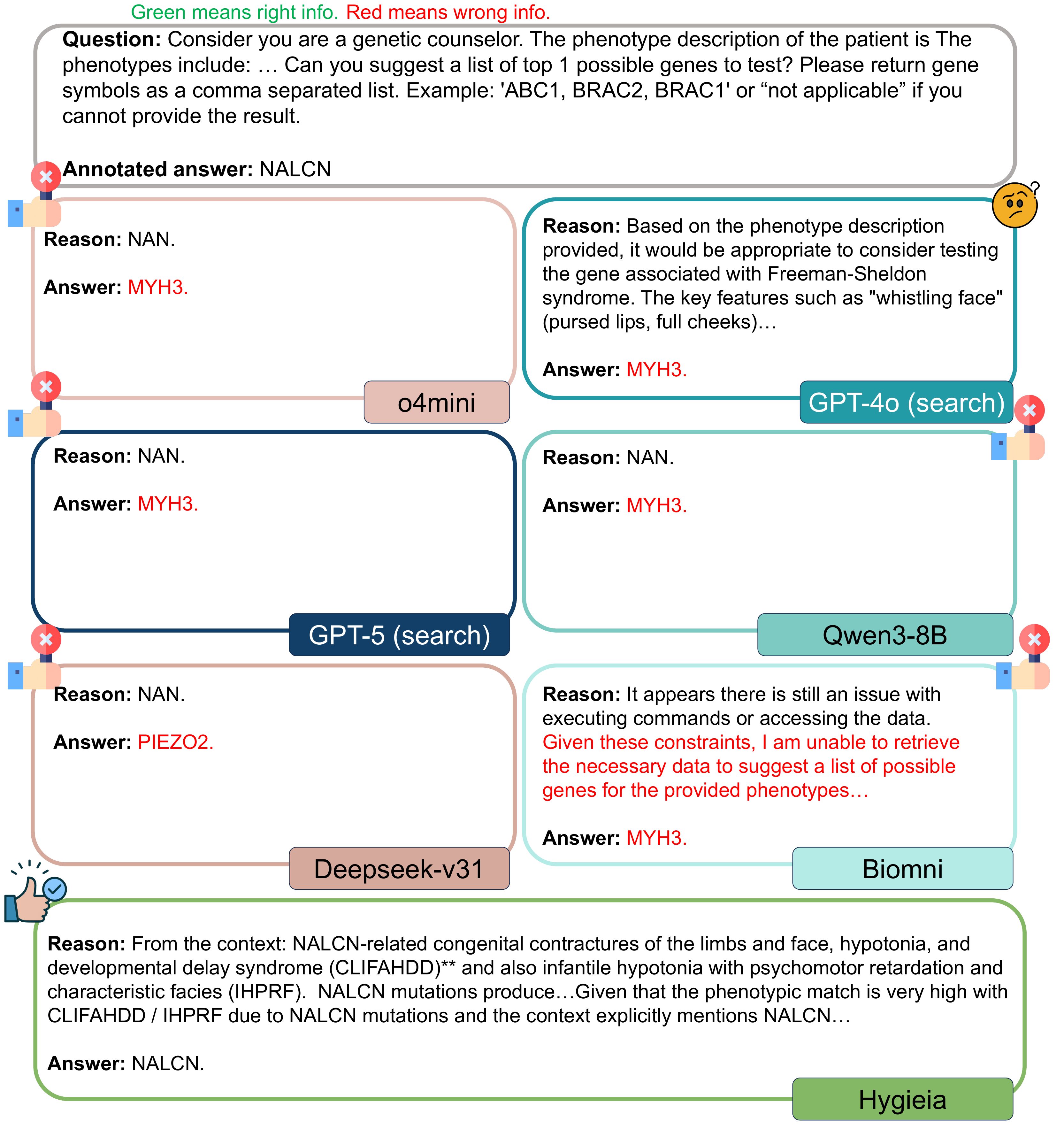}
    \caption{Case study for \method{} and other baselines in risk gene prioritization. We mask some phenotypes to protect personal information.}
    \label{fig:gene_example_case}
\end{figure}

According to Figure \ref{fig:gene_example_case}, in this evaluation, \method{} is the only system that successfully identifies NALCN, which is the correct gene associated with the patient’s constellation of phenotypes, fully aligning with the annotated ground truth. All other baseline models, including o4mini, GPT-4o (search), GPT-5 (search), Qwen3-8B, and DeepSeek-v31, either returned incorrect candidates such as \textit{MYH3} or \textit{PIEZO2}, or failed to provide a usable response altogether. This example demonstrates that \method{} consistently outperforms both general-purpose LLMs and search-augmented tools in high-stakes biomedical inference tasks requiring precise gene–phenotype matching.

Unlike competing models that default to \textit{MYH3} based on superficial resemblance to Freeman–Sheldon syndrome, \method{} distinguishes clinically overlapping yet genetically distinct disorders by integrating phenotypic, molecular, and nosological evidence. The model explicitly links the observed features—including congenital contractures, hypotonia, neurodevelopmental delay, and characteristic facies, which are NALCN-associated CLIFAHDD/IHPRF syndromes. This indicates that \method{} does not rely solely on phenotypic pattern matching, but instead performs multi-layered biomedical reasoning consistent with genetic counseling practice.

Several baselines fail due to inability to retrieve or interpret data, producing unusable outputs (``NAN", ``unable to retrieve data"). \method{} remains fully functional even under incomplete signal, reflecting robustness to real-world clinical constraints, where patient phenotypes may be sparsely documented, noisy, or partially overlapping. This resilience is essential for deployment in clinical decision support, where diagnostic completeness is rarely guaranteed.

\method{} not only outputs the correct target gene but also contextualizes it within a clinically actionable diagnostic category. This stands in sharp contrast to competing models that provide unsubstantiated gene names without justification, which would be unacceptable in a clinical genetics workflow where gene testing decisions have financial, ethical, and prognostic implications. \method{}’s interpretability and biological validity, therefore, make it a more trustworthy candidate for integration into precision medicine pipelines.

The comparison clearly illustrates that \method{} surpasses existing LLM-based and search-augmented baselines in terms of accuracy, reasoning validity, and clinical relevance. Its ability to discriminate among phenotypically similar developmental syndromes and return a gene with direct translational value underscores its potential as a next-generation AI system for genetic factor prioritization.

\textbf{Utilizing \method{} as a medical assistant for physicians in solving complicated cases.} One major goal of developing and deploying \method{} is to make a virtual assistant (Copilot) for physicians and clinicians working on rare disease diagnosis and treatment development, and thus, matching user requirements with the functionality of \method{} is a crucial step. Previous medical AI agent development has not explored this area extensively, thereby limiting their specific deployment capabilities. In this work, we collaborated with physicians from Duke-NUS, YSM, and YNHH to define key stakeholders in disease diagnosis/gene prioritization, and chart a blueprint for human-AI collaboration, thereby providing a guidance framework for the concrete implementation of \method{}.

\begin{figure}[H]
    \centering
    \includegraphics[width=1\linewidth]{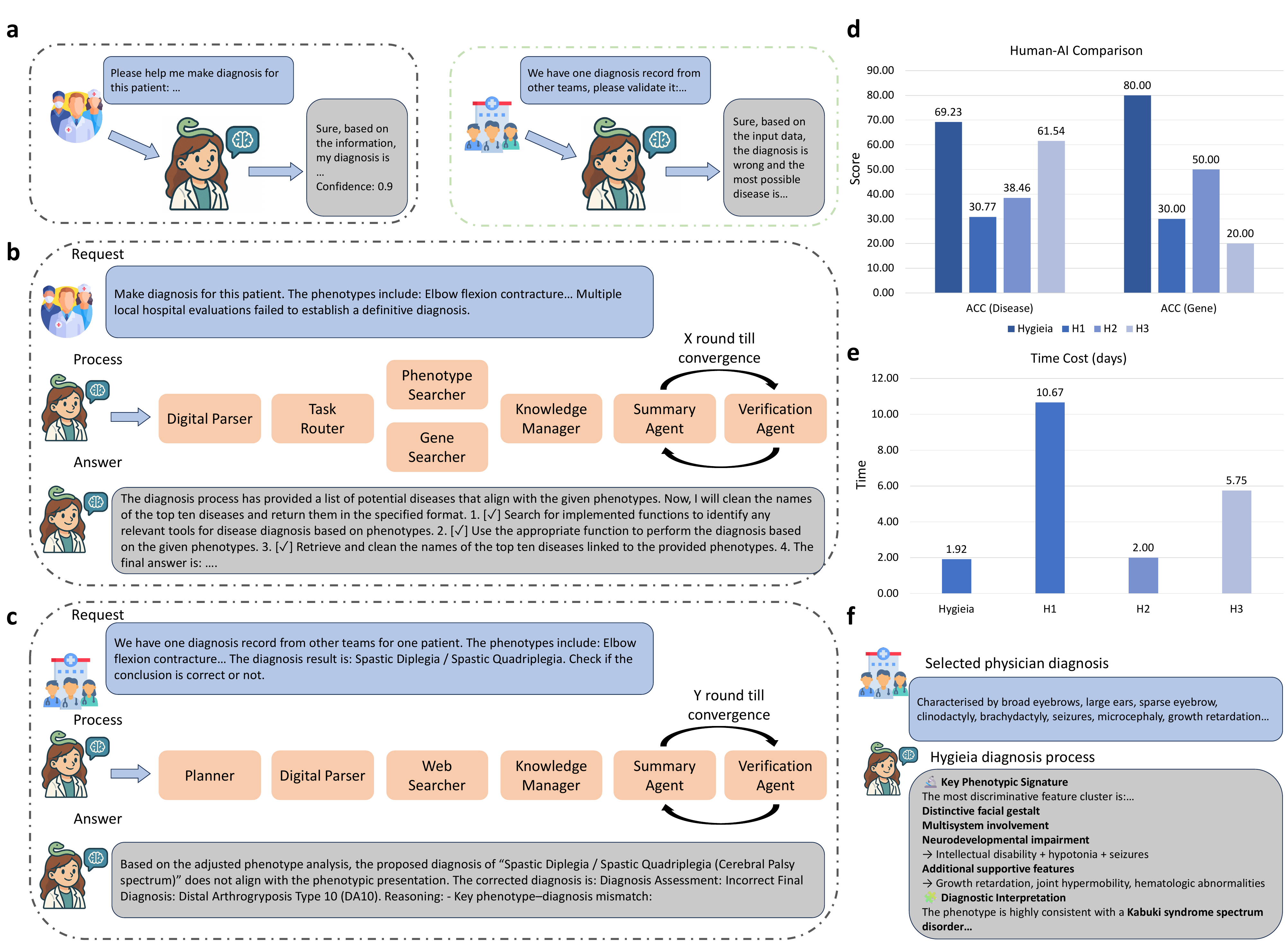}
    \caption{Illustration of Human-AI collaboration for disease diagnosis and decision correction based on physicians and \method{}. We mask phenotype information to protect patients' privacy. (a) Explanations of two selected tasks; (b) Illustration of the diagnosis of diseases based on physician input; (c) Illustration of verification and refinement of physician judgments. (d) Human-AI comparison results for two tasks, H1-H3, represent human experts. (e) Time cost comparison result. (f) Case study for outcomes from our agent and human experts.}
    \label{fig:examples_humaneval}
\end{figure}

Figure \ref{fig:examples_humaneval} (a) shows two expected functions of \method{} from physicians to improve the efficiency and reduce the effort, including direct diagnosis based on physician input, and verification and refinement of physician judgments. The commonality between these two tasks is that both require interaction between the physician and the AI agent, where the physician articulates the character's needs, and the agent fulfills them. We also provide some examples here to showcase how we can use \method{} to accomplish these tasks. In Figure \ref{fig:examples_humaneval} (b), we illustrate that \method{} can take the physicians' instructions with patient phenotype information (free text format) and other relevant information as inputs, and integrate its different components to create a pipeline, call different tools, and communicate with different AI agents to produce a diagnosis result. In Figure \ref{fig:examples_humaneval} (c), we showcase how \method{} can take the physicians' diagnosis results as well as phenotype information as inputs, and by using an alternative best pipeline, first determine the True/False or original diagnosis, and then perform reasoning and correction, to report a new diagnosis answer. 

We also invite three \textbf{certified genetic physicians} as experts in pediatrics and rare disease diagnosis to have a competition with \method{} based on the two selected tasks discussed in this manuscript. Their training backgrounds are summarized in Supplementary File 1 to justify their strong expertise. We assigned the same question sets for human experts and \method{}, allowing them to access the same resources (but human experts are not allowed to use LLMs as assistants), and make a direct comparison in their abilities for problem solving. According to Figure \ref{fig:examples_humaneval} (d), \method{} performs better than all selected human experts in both rare disease diagnosis and gene prioritization, and the improvements versus the second-best expert are 12.49\% and 60\%, respectively. Human experts also have comments that gene prioritization is more difficult than disease diagnosis, supported by the difference in the improvement. Regarding the efficiency, \method{} only spends less than two hours in solving all of the selected questions, while human experts take from two hours to 10 hours to answer all questions, shown in Figure \ref{fig:examples_humaneval} (e). Therefore, AI-based medical assistants are able to provide feedback more quickly; this capacity could be very valuable, especially in cases related to timely decisions. Finally, we show a case study for the previously mentioned challenging case, the Kabuki syndrome, and analyze the reasons from human experts and \method{} to make a diagnosis. According to Figure \ref{fig:examples_humaneval} (f), human experts prefer giving reasons with conciseness, and selecting highly-relevant phenotypes to make a diagnosis. However, \method{} prefers generating a more structured reasoning path with hierarchy and boldfacing the important information. Therefore, \method{} can facilitate reasoning for disease diagnosis with an easier approach to understanding and analysis, especially for junior physicians and medical students learning the diagnosis process.

The robust performance of \method{}, reflected in both its reasoning process and diagnostic outputs, together with its advantage over human experts in comparative evaluations demonstrates its potential to translate into tangible medical value for healthcare teams, further highlighting its dual contributions at both the algorithmic and application levels. Moving forward, we will involve physicians as human evaluators to directly compare \method{} with physicians in tasks such as disease diagnosis and key gene identification, thus further extending \method{}.

%% file: section_folder/discussion.tex
Accurate and timely diagnosis remains a fundamental challenge in modern medicine, particularly for rare and genetically heterogeneous diseases where phenotypic overlap, incomplete records, and limited clinician exposure often result in prolonged diagnostic odysseys and misdiagnoses. While recent advances in LLMs and biomedical AI have demonstrated promise in clinical reasoning, most existing approaches focus narrowly on either pattern-based disease prediction or retrieval-augmented question answering, without addressing the full diagnostic workflow, interpretability, or clinical deployment constraints. In this work, we present \method{}, a versatile multi-modal agentic system designed to jointly address disease diagnosis and risk gene prioritization by integrating phenotypic, genomic, and contextual clinical information within a unified framework powered by AI.

Overall, \method{} introduces several conceptual and technical advances that collectively address the above challenges. First, it formalizes diagnosis as a multi-stage agentic workflow, beginning with task parsing and disease-type routing, followed by tailored pipelines for common and rare diseases. By explicitly separating these pathways, \method{} avoids over-generalized reasoning and reduces both computational cost and diagnostic error for common conditions, while allocating more sophisticated reasoning resources to rare disease cases. Second, the system incorporates a self-verification mechanism that iteratively evaluates and corrects intermediate diagnostic outputs until convergence. This verifier–corrector design substantially improves consistency and robustness, mitigating the well-documented randomness of LLM-based inference. Empirically, this design leads to more stable predictions, and we also provide better-calibrated confidence estimates with multiple queries. Third, \method{} extends beyond diagnosis to risk gene prioritization, framing it as a closely related inference task that benefits from the same phenotypic and contextual reasoning. By unifying these tasks in a single framework, we produce clinically actionable outputs that can directly empower several associated tasks. Fourth, \method{} also emphasizes interpretability and clinical alignment. Rather than producing opaque predictions or generic summaries, our system explicitly links phenotypes, disease entities, and gene-level evidence through multi-step reasoning trajectories. 

Moreover, \method{} is evaluated not only through comprehensive benchmarks across diverse datasets with imbalanced and noisy data as inputs, but also through human-in-the-loop studies with practicing physicians, demonstrating its utility as a diagnostic assistant and verifier of clinician judgments. We not only test the performances of \method{} versus human experts in selected tasks, but also showcase several case studies to demonstrate the contribution of Human-AI collaboration. This dual evaluation strategy strengthens the translational relevance of the proposed system.

Despite its strong performance, \method{} has several limitations that warrant discussion. First, it relies heavily on the capabilities of underlying LLMs, many of which are closed-source and subject to evolving behaviors, cost structures, and access constraints. Second, while \method{} demonstrates strong generalization across datasets, its evaluation is still constrained by the availability and quality of labeled rare disease data, particularly for gene-level ground truth. Simulation datasets are not enough to directly validate their real-world applications in more complicated scenarios. In the future, we will focus on extending \method{} via more flexible model selection as well as constructing more datasets for the agentic system training and validation.

%% file: section_folder/methods.tex
\textbf{Problem definition.} Here we intend to design a specific agentic system for disease-relevant analysis, including disease diagnosis and risk factor (such as gene) prioritization. Our agent system $\mathcal{A}$ can take the following information as inputs, including task-specific prompts $R\_x$, phenotypes $P\_x$, functional and/or genetic information of risked genes $G$, and medical record information $C$. For the problem of disease diagnosis, this system makes inference with $O_i=\mathcal{A}(R_d, P_i, G_i, C_i)$ for the patient $i$, and the diagnosis result will be the output. Similarly, for risk gene prioritization, we replace the task-specific prompt $R_d$ with $R_g$, and the output will be the prioritized gene. The model output not only contains the expected diseases or genes (in string), but also the confidence of making such a decision ranging from 0-100. 

\textbf{Workflow of \method{}.} Our workflow contains four main stages, including \textit{task-specific planning}, \textit{information retrieval and integration}, \textit{self-reflection-based validation}, and \textit{confidence estimation}. By default, most agents are implemented using GPT-5 \cite{openai2025gpt5} as their backbone after considering the trade-off among model performance, cost, and protection of patient privacy. We use Claude-Sonnet-4.5 \cite{anthropic2025claudesonnet45} as the backbone of our verification agent to reduce model bias. We have discussed the ablation studies in the Methods section to justify our settings.

Regarding \textit{task-specific planning}, we can integrate information proposed by known biomedical databases and divide the main task into several steps, based on Biomni \cite{huang2025biomni}. The agent will parse the input information first, and search the current tool base developed based on both tools in Biomni and newly implemented searching functions. It will then determine the correct tool for addressing the given task. We do not map phenotypes with HPO terms \cite{robinson2010human} as we assume that our agent (based on advanced LLMs) already knows related information. Regarding the difference in diagnosing common and rare diseases, we train a classifier-based router to make more precise diagnoses and reduce cost based on a KNN classifier \cite{pedregosa2011scikit}. This stage is performed by the parser and the router component.

Regarding \textit{information retrieval and integration}, we search the related information of phenotypes and possible gene names as well as gene functions based on web-searching tools, and the data sources including Google, Google Scholar, and PubMed \cite{white2020pubmed}. Specific human variants and related information are not included in the current version due to the protection of privacy. We also allow the advanced searching tool in LLMs such as the web-searching tools in GPT-4o \cite{hurst2024gpt} and GPT-5 \cite{openai2025gpt5}. Here, we consider searching the top $K=5$ patients from known databases with diagnosis information as references. After collecting the necessary information, we will integrate the prior knowledge as inputs for the next component in this system. This task is finished by the knowledge-manager, web-searcher, and patient-retriever components.

Regarding \textit{self-reflection-based validation}, we provide the diagnosis decision as well as methods for validation. The summary agent in our system will utilize the information from the previous two stages and generate a clinical decision. After making the decision based on the summary agent, we introduce our verification agent, which takes the prior knowledge and the output of the summary agent as input and validates whether the result is correct or incorrect \cite{yao2022react}. If the result is incorrect, the prompt used for the summary agent will update, and the summary agent will make a decision again, until the verification agent agrees with the decision or we reach the limit of tries. The algorithm of this system is shown in Algorithm 1, with the disease diagnosis task as an example. This task is finished by the summary-agent and verification-agent components. 

\begin{algorithm}[htbp]
\caption{Verifier-Corrector Pipeline in \method{} for disease diagnosis.}
\begin{algorithmic}[1]
\Statex \textbf{Input:} Question $Q_A$, verify prompt $T_V$, model prompt $T_C$, number of iteration $N$, knowledge query prompt $T_K$.
\Statex \textbf{Helper Models:} Verifier $\mathcal{M}_v$, diagnosis agent $\mathcal{M}_s$, router $\mathcal{M}_r$, knowledge extractor $\mathcal{M}_k$, knowledge manager $\mathcal{M}_m$, concatenation function $\cdot||\cdot$.
\Statex \textbf{Intermediate variable:} Model output $O_A$, reasoning path $O_R$, external knowledge $K_A$.
\Statex \textbf{Output:} Verified model output $O_C$, reasoning output $R_A$.
\State INIT: initialize all parameters.
\If{$\mathcal{M}_r(Q_A)$ is Common}
\State $O_C,R_A=\mathcal{M}_s(T_C,Q_A)$
\State Return $O_C,R_A$
\EndIf
\State $K_A=\mathcal{M}_k(T_K,Q_A)$
\State $O_K=\mathcal{M}_m(T_K,K_A)$
\State $O_A,R_A=\mathcal{M}_s(T_C,Q_A||O_K)$
\For{$i$ in $N$ steps}
\If{$\mathcal{M}_v(T_V,Q_A||O_A)$ is True}
    \State $O_C=O_A$
    \State Return $O_C,R_A$
\Else
    \State $O_A,R_A=\mathcal{M}_s(T_C,Q_A||O_K)$
\EndIf
\EndFor
\State $O_C,R_A=\mathcal{M}_s(T_C,Q_A||O_K)$
\State Return $O_C,R_A$
\end{algorithmic}
\end{algorithm}

Regarding \textit{confidence estimation}, we refer to the method introduced by \cite{junxianhecanllm}, and we ask the summary agent $s$ times to get $s$ answers as well as $s$ paired confidence lists $c_1,c_2,...,c_s$. We average the confidence levels and use the major voting result from these $s$ answers as the final decision. Therefore, the final confidence is $c_f = \frac{\sum_{k=1}^s c_k}{s}$. We have tried other methods, including summarization of logprobs, self-evaluation \cite{kadavath2022language}, and thinking-twice-before-answering \cite{li2024think}, but their performances are not good enough to represent the confidence. Since most of the advanced LLMs are closed-source and black-box models, other approaches used for open-source LLMs are not applicable. We have performed statistical tests to demonstrate that our current settings can help us calibrate the model outputs.

\textbf{Ablation studies.} In Supplementary Figures \ref{supfig:abla_study} (a) and (b), we show the contributions of adjusting the base models and the input information for these tasks. In both cases, changing the base models from GPT-4o to GPT-5 makes an obvious improvement, and incorporating more context, such as detailed phenotype descriptions, as well as having a verifier, can help \method{} determine diseases and gene sets. We also find that providing disease information can help \method{} rank genes; however, since in real clinical cases, physicians must know the results of the genetic tests and then make a diagnosis, we do not use this information as input for the prioritization of risk genes. 

\textbf{Human-in-the-loop design.} We also consider the comparison and collaboration between physicians and AI models in making decisions. We have invited four genetic physicians from YSM, YNHH, and Duke-NUS Medical School, and consider two scenarios. 

In our first experimental setting, we assign 23 questions (13 for disease diagnosis with 2 difficult questions related to extremely rare diseases, and 10 for gene prioritization) for X physicians and \method{}, and evaluate the performances based on accuracy. 

In the second experimental setting, we allow \method{} to access the results provided by physicians, and utilize \method{} for verifying the decision from physicians. If the verification turns out with negative results, we will run the correction pipeline discussed in our Methods and Results section to update the decision.



\textbf{Case study investigation.} To better demonstrate the advantages of \method{}, we provide several case studies from different baseline methods and include the reasoning steps of \method{}. In these questions, only \method{} makes the correct decision, while the rest of the methods do not give us either correct reasoning paths or correct answers.

\textbf{Explanations of baseline methods.} For closed-source models, we consider LLMs with reasoning capacities including o4-mini \cite{o3o4mini-systemcard-2025}, GPT-5 \cite{openai2025gpt5}, and GPT-5 (search) \cite{openai2025gpt5}, advanced LLMs including GPT-4o \cite{hurst2024gpt} and GPT-4o (search) \cite{hurst2024gpt}. For open-source models, we consider Qwen3-8B \cite{yang2025qwen3} and Deepseek-v3.1 \cite{liu2024deepseek}. For domain-expert models, we consider a multi-agent diagnosis system (base model GPT-4o as the recommended setting), as well as DeepRare \cite{zhao2025agentic}. However, DeepRare's API version has bugs related to data processing, and we cannot find their released testing datasets, making it hard for us to retrieve the data format. For biomedical agents, we consider Biomni (base model GPT-4o as the recommended setting. For Qwen3-8B, we also consider fine-tuning the base model with both Low-rank adaptation (LoRA) and full parameters (Full) \cite{zheng2024llamafactory} based on the provided simulation data. The prompts used to query LRM and LLMs are documented in Appendix \ref{appendix:prompt}. For other models, details can be found in our code base.

\textbf{Metrics.} We follow the settings discussed in \cite{chen2024rarebench,zhao2025agentic}, where the Recall$@$K is the main metric used in evaluating the generated disease diagnosis and gene rank results. We consider the top 1,5,10 candidates and compute Recall$@$1, Recall$@$5, Recall$@$10, accordingly. We perform metric computation after disease name/gene name normalization.

\section{Code Availability and Data Availability}
We use the authorized OpenAI API, Claude API, and Gemini API to develop our method and perform benchmark studies. To fine-tune open-source LLMs, we use the Misha cluster from Wu Tsai Institute at Yale with one NVIDIA H100 GPU and 80GB of memory. The codes of \method{} can be found in \url{https://github.com/HelloWorldLTY/hygieia} with the MIT license. 

All data are available online or upon request. We have summarized the download links of each dataset in Supplementary File 2. 

\section{Institutional Review Board (IRB) Approval}
This project has received approval from Yale IRB, with project number 2000039055. 

\section{Acknowledgment}
We acknowledge the support from the Undiagnosed Diseases Network. T.L. acknowledges the support from the OpenAI Research Access Program. 

\section{Author Contribution}
T.L. designed this study with W.Z. and H.Z. T.L. developed the method. T.L. performed experiments. B.K.G.K., H.Z., and J.L. performed human evaluations. All authors contributed to manuscript writing and reviewing. H.Z. supervised this project.

%% file: section_folder/appendix.tex
\counterwithin{table}{section}
\renewcommand{\tablename}{Supplementary Tab.}
\renewcommand\thetable{\arabic{table}}  
\section{Comparison between \method{} and other AI agents}
\label{appendix:agentcompare}

\begin{table*}[h]
\centering
\small
\resizebox{0.98\textwidth}{!}{
\begin{tabular}{lcccccccc}
\toprule
\textbf{System} & \textbf{Distinguish} & \textbf{Human} & \textbf{Confidence} & \textbf{Risk Gene} &  \textbf{Flexible Input} &  \textbf{Open Source} &  \textbf{Evaluation Setting} & \textbf{Domain} \\

\midrule

MCA \cite{chen2025enhancing}& \xmark & \xmark & \xmark& \xmark & \cmark & \cmark & Benchmark & Rare \\
MDAgent \cite{kim2024mdagents}& \cmark & \xmark & \xmark& \xmark & \cmark & \cmark & Benchmark & General \\
DeepRare \cite{zhao2025agentic}& \xmark & \xmark & \xmark& \xmark & \xmark & \cmark  & Benchmark & Rare \\
Biomni \cite{huang2025biomni}& \cmark & \xmark & \xmark& \cmark & \cmark & \cmark & Benchmark \& Human Eval & General \\
RDguru \cite{yang2025rdguru}  & \xmark & \xmark & \xmark& \xmark & \xmark & \xmark & Benchmark & Rare \\
RareAgent \cite{chen2024rareagents}  & \xmark & \xmark & \xmark& \xmark & \xmark & \xmark & Benchmark & Rare \\
\midrule
\method{} (Ours) & \cmark & \cmark & \cmark & \cmark & \cmark & \cmark &  Benchmark \& Human Eval & Rare \\
\bottomrule
\end{tabular}
}
\caption{
Comparison of representative AI Agents for disease-related tasks across different dimensions. Here we consider Distinguish (whether the agent can distinguish common and rare diseases), Human (whether the agent supports human-in-the-loop), Confidence (whether the agent can produce confidence), Risk Gene (whether the agent can also infer risk gene), Flexible Input (whether the agent supports input other than HPO terms), and Open Source (whether the codes are accessible). We also compare their evaluation settings as well as focused domains.
}
\label{tab:agent_comparison}
\end{table*}

In Supplementary Table \ref{tab:agent_comparison}, we showcase the unique contribution of \method{} by comparing it with other disease diagnosis agents across multiple dimensions.

\section{Prompts}
\label{appendix:prompt}
The prompts used for disease diagnosis: ``Make diagnosis for this patient. Known phenotypes include: \{phenotype\_list\}. Multiple local hospital evaluations failed to establish a definitive diagnosis."

The prompts used for gene prioritization: ``Consider you are a genetic counselor. The phenotype description of the patient is \{phenotype\_list\}. Can you suggest a list of top 1 possible genes to test?"

The prompts used for error detection and correction:

You are a board-certified clinical geneticist and neurologist with expertise in rare neuromuscular and congenital disorders.
You reason step-by-step using established diagnostic criteria, genotype–phenotype correlations, and differential diagnosis logic.

Below is a patient’s clinical phenotype and a proposed diagnosis.

Your task is to determine whether the proposed diagnosis is correct.

Instructions:
1. Carefully assess whether the phenotype is consistent with the proposed diagnosis.
2. If the diagnosis is correct, explicitly state that it is correct and explain why.
3. If the diagnosis is incorrect or incomplete, clearly state that it is incorrect and:
   - Provide the most likely corrected diagnosis
   - Briefly justify the correction using key phenotype–disease matches
4. Do not provide multiple diagnoses—return one best diagnosis only.
5. Be concise, clinically precise, and avoid speculation beyond the given phenotype.

Patient Phenotype:
\{PHENOTYPE\_LIST\}

Proposed Diagnosis:
\{PROPOSED\_DIAGNOSIS\}

Output Format (strict):

Diagnosis Assessment: Correct / Incorrect

Final Diagnosis: <single diagnosis name>

Reasoning:
- Key phenotype–diagnosis alignment (or mismatch)
- Critical features supporting the final diagnosis

\newpage

\counterwithin{figure}{section}
\renewcommand{\figurename}{Supplementary Fig.}
\renewcommand\thefigure{\arabic{figure}}  

\section{Supplementary Figures}
\begin{figure}[H]
    \centering
    \includegraphics[width=0.9\linewidth]{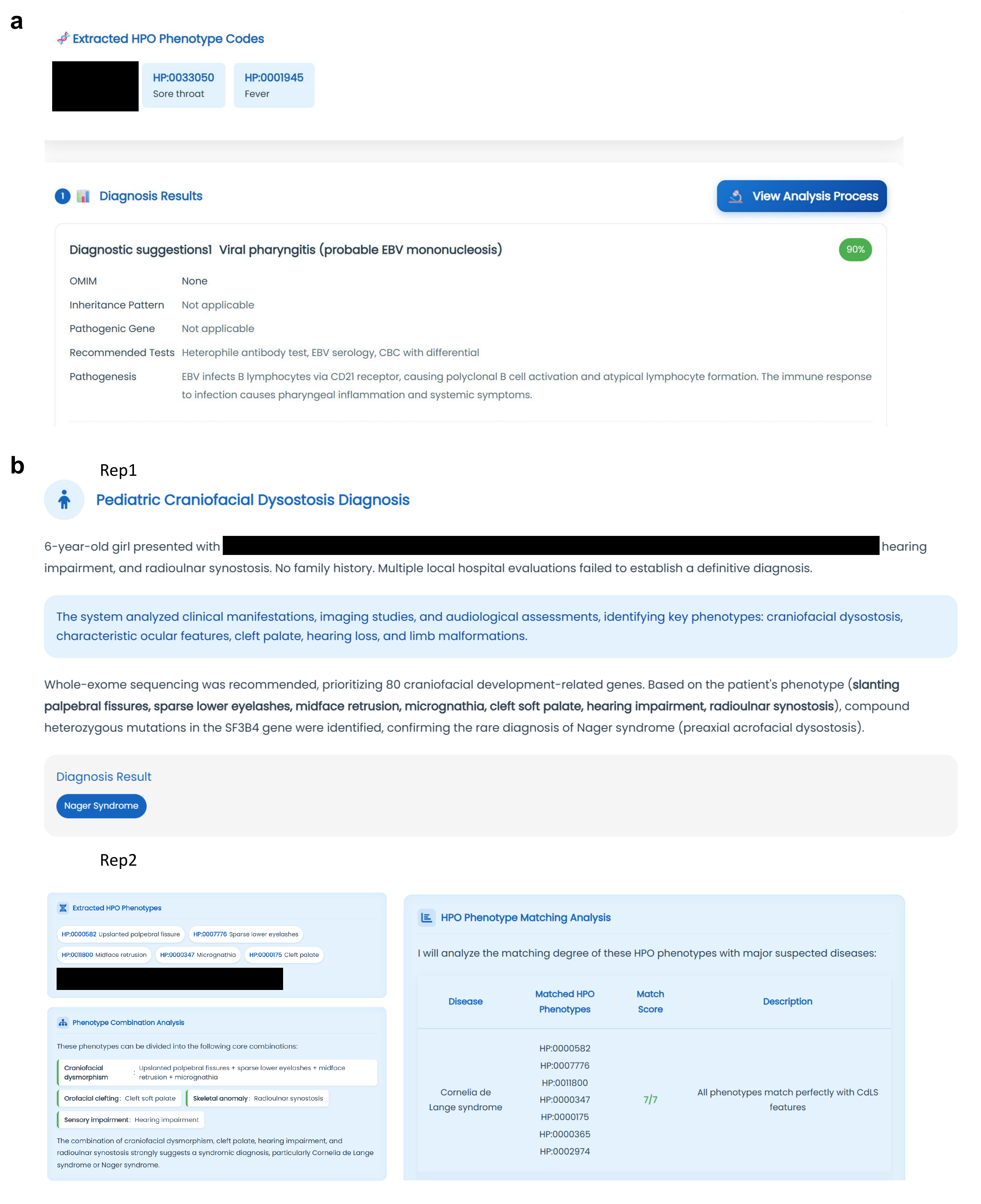}
    \caption{Failure cases of AI agent in disease diagnosis, by using DeepRare as an example. We mask some phenotypes to protect patient information. (a) Failure of diagnosis. (b) Failure of reproducing results.}
    \label{supfig:deeprare_case}
\end{figure}

\begin{figure}
    \centering
    \includegraphics[width=1\linewidth]{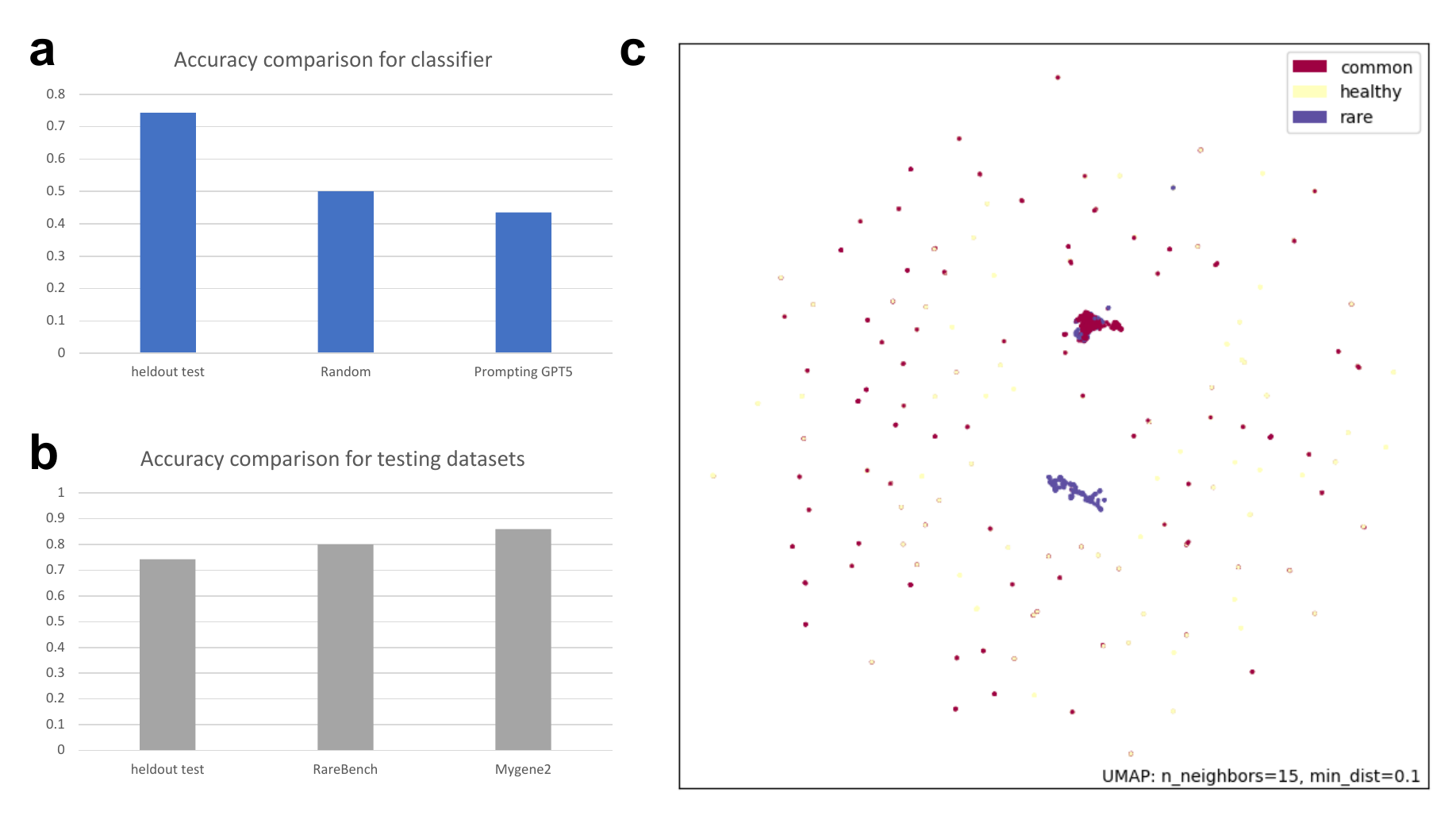}
    \caption{Classification results of router. (a) Accuracy comparisons of different classification methods. (b) Accuracy comparisons of different testing sets. (c) UMAP colored by sample labels (common diseases, rare diseases, and healthy people).}
    \label{supfig:diseasrouter}
\end{figure}

\clearpage

\begin{figure}
    \centering
    \includegraphics[width=1\linewidth]{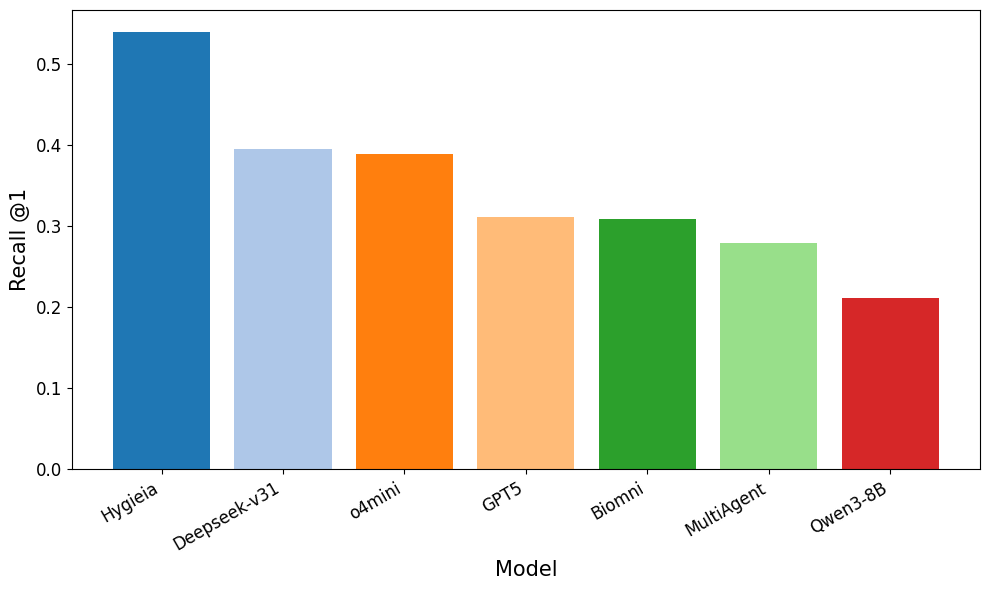}
    \caption{Benchmarking results for disease diagnosis based on the RareArena dataset.}
    \label{supfig:rarearena_info}
\end{figure}

\clearpage

\begin{figure}
    \centering
    \includegraphics[width=1\linewidth]{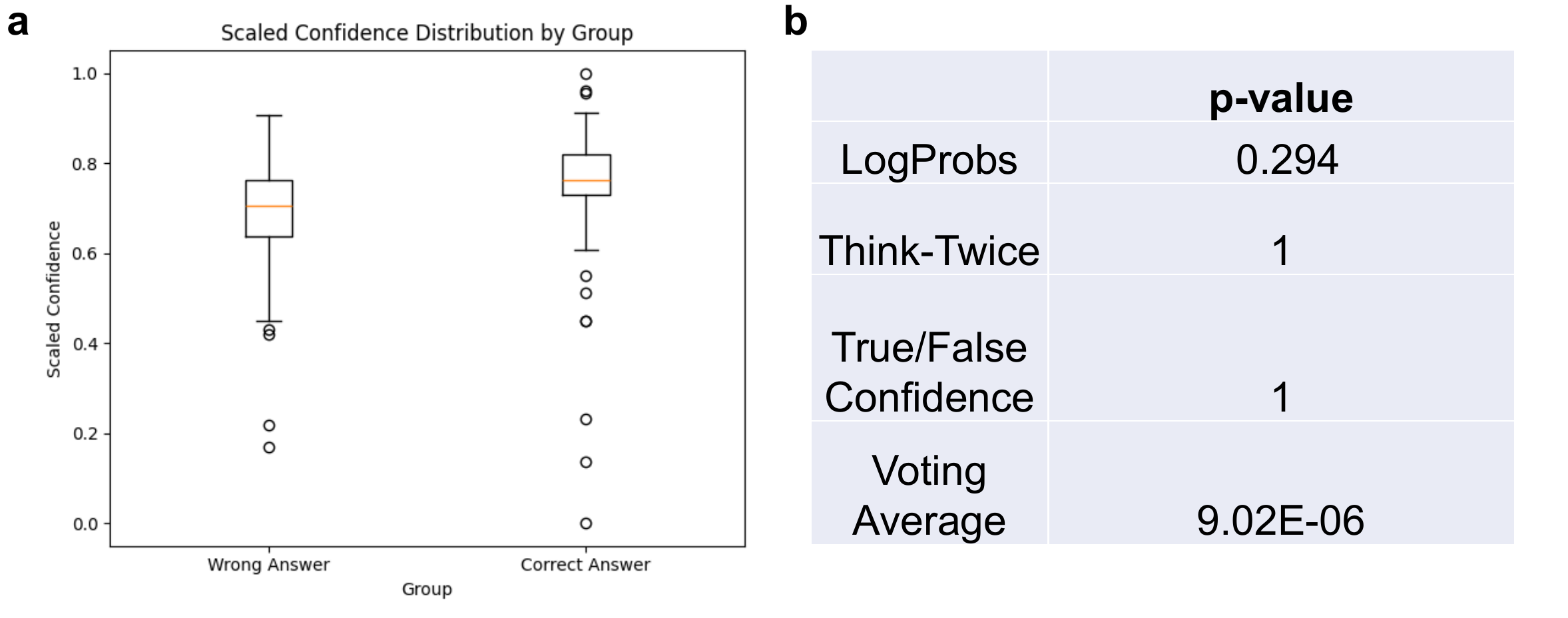}
    \caption{Estimation of confidence levels based on \method{}. (a) Comparison of scaled confidence scores tested with MyGene2 dataset. (b) Ablation studies for different estimation methods. The p-value is computed based on two-sided Mann-Whitney U test. }
    \label{supfig:conf_est}
\end{figure}

\clearpage

\begin{figure}
    \centering
    \includegraphics[width=1\linewidth]{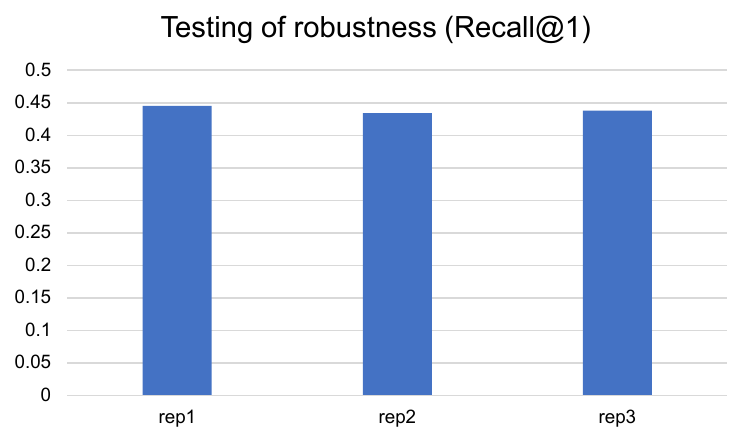}
    \caption{Examination of AI agent robustness.}
    \label{supfig:robustness_check}
\end{figure}

\clearpage

\begin{figure}
    \centering
    \includegraphics[width=1\linewidth]{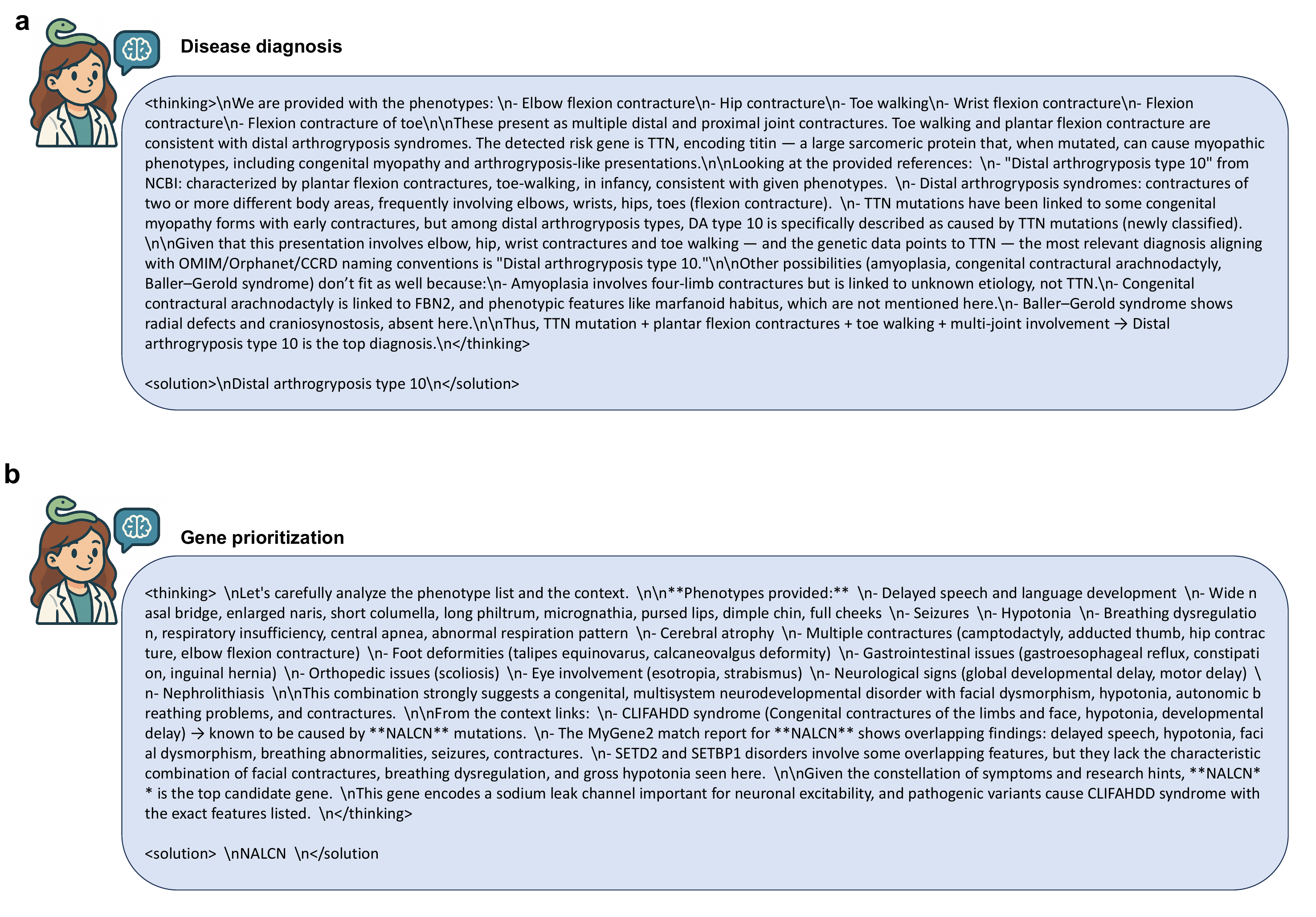}
    \caption{Full reasoning paths in two tasks by \method{}. (a) is for disease diagnosis and (b) is for risk gene prioritization.}
    \label{supfig:raredis_example}
\end{figure}

\clearpage

\begin{figure}
    \centering
    \includegraphics[width=1\linewidth]{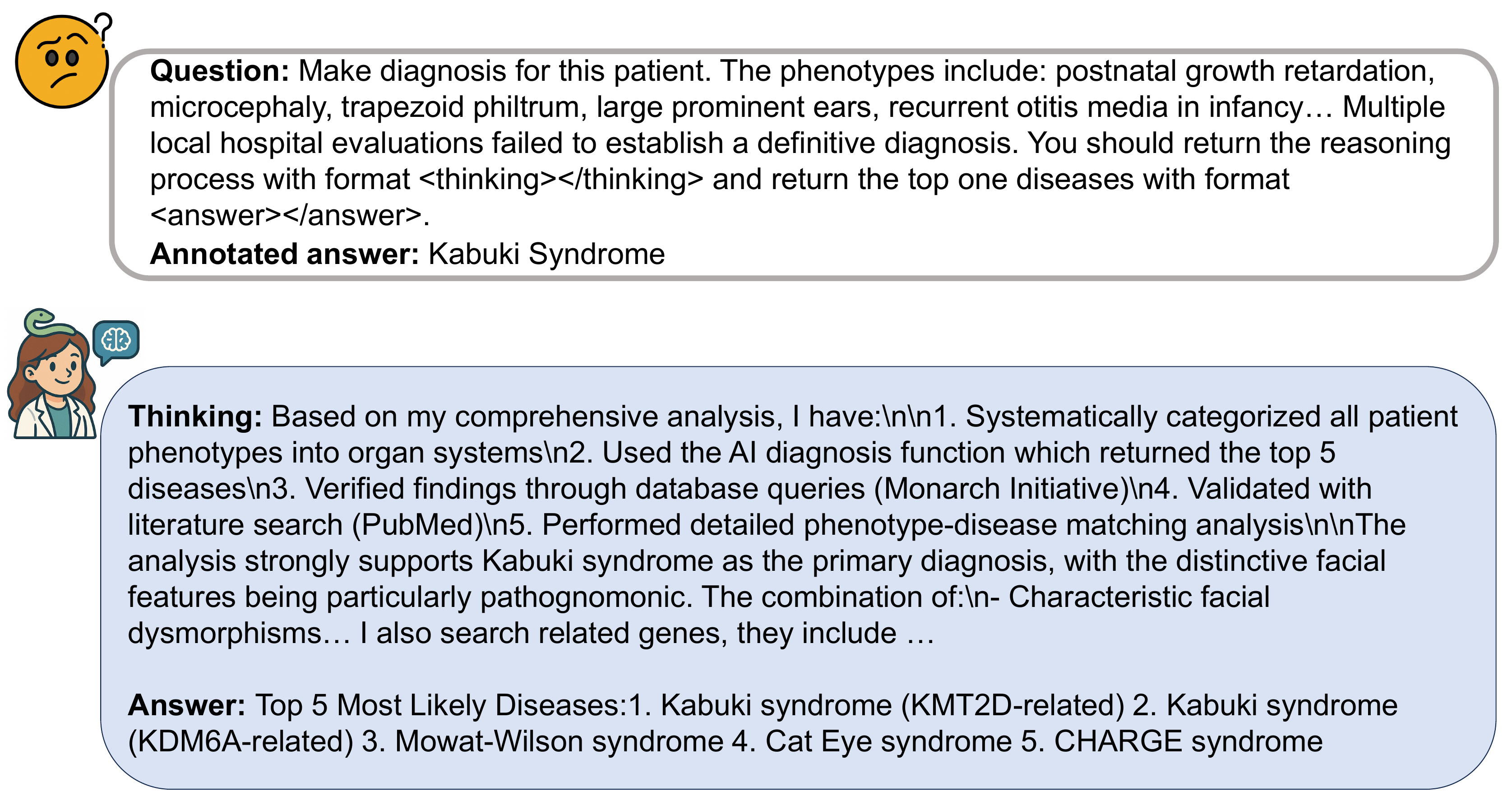}
    \caption{Questions and answers from \method{} based on a difficult selected sample from our in-house data.}
    \label{supfig:raredis_diff_example}
\end{figure}

\clearpage

\begin{figure}
    \centering
    \includegraphics[width=1\linewidth]{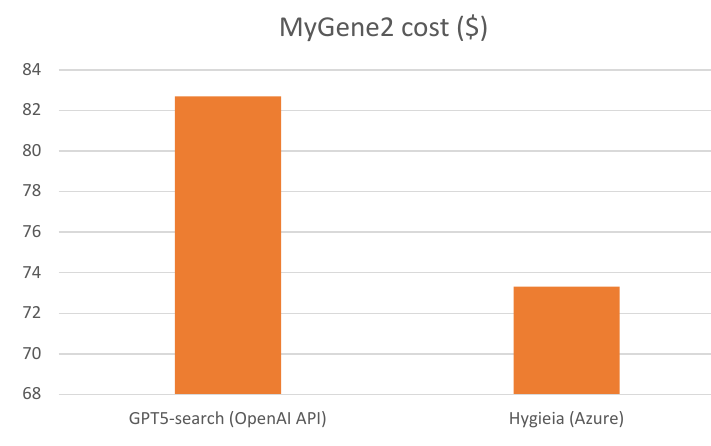}
    \caption{Cost analysis between GPT-5 Search and \method{} in risk gene prioritization based on MyGene2.}
    \label{supfig:costanalysis}
\end{figure}

\begin{figure}
    \centering
    \includegraphics[width=1\linewidth]{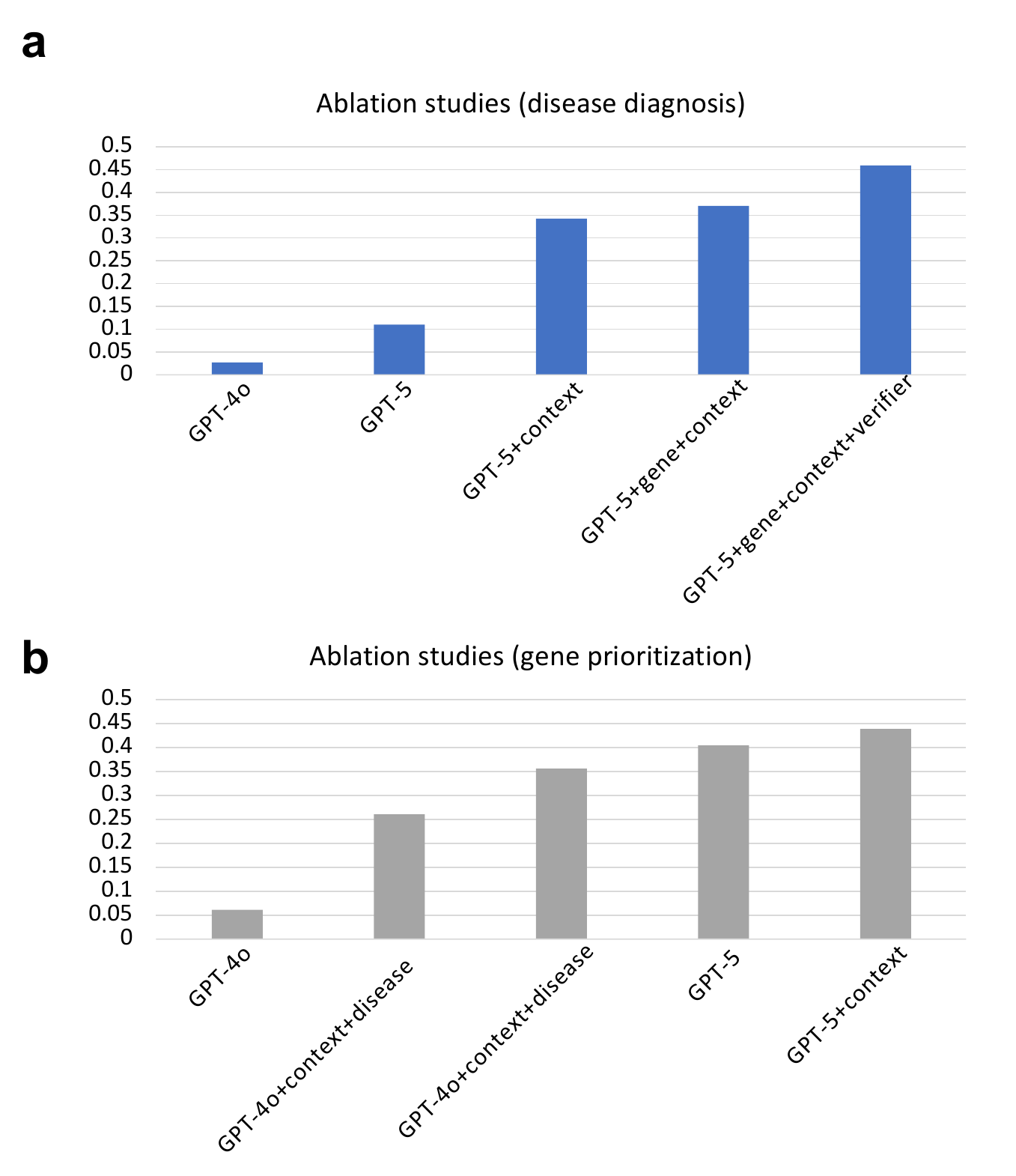}
    \caption{Ablation studies of \method{}. (a) represents the results for disease diagnosis and (b) represents the results for risk gene prioritization.}
    \label{supfig:abla_study}
\end{figure}